\documentclass[10pt,twocolumn,letterpaper]{article}

\usepackage{iccv}
\usepackage{times}
\usepackage{epsfig}
\usepackage{graphicx}
\usepackage{amsmath}
\usepackage{amssymb}
\usepackage{multirow}
\usepackage{caption}
\usepackage[table,x11names]{xcolor}
\usepackage{soul}

\graphicspath{ {Ims/} }

\usepackage[pagebackref=true,breaklinks=true,letterpaper=true,colorlinks,bookmarks=false]{hyperref}

\newcommand{\ignorethis } [1] {}

\newcommand{\eqnnum     } [1] {\mbox{(\ref{#1})}}

\newcommand{\eqn        } [1] {Equation~\eqnnum{#1}}

\iccvfinalcopy 

\def\iccvPaperID{2707} 

\ificcvfinal\fi
\begin{document}

\title{Training Deep Networks to be Spatially Sensitive}

\author{\hspace{1cm}Nicholas Kolkin \hspace{0.95cm}Gregory Shakhnarovich\\
{\tt\small nick.kolkin@ttic.edu}\hspace{1.45cm}{\tt\small gregory@ttic.edu}\\
Toyota Technological Institute at Chicago\\
 Chicago, IL, USA
\and
Eli Shechtman\\
{\tt\small elishe@adobe.com}\\
Adobe Research\\
Seattle, WA, USA
}
\maketitle
\thispagestyle{empty}

\begin{abstract}
    In many computer vision tasks, for example saliency prediction or
    semantic segmentation, the desired output is a foreground map
    that predicts pixels where some criteria is satisfied. Despite
    the inherently spatial nature of this task commonly used learning objectives do not incorporate the spatial relationships 
    between misclassified pixels and the underlying ground truth. The Weighted F-measure, a recently proposed evaluation metric, does reweight errors spatially, and has been shown to closely correlate with human evaluation of quality, and stably rank predictions with respect to noisy ground truths (such as a sloppy human annotator might generate). However it suffers
    from computational complexity which
    makes it intractable as an optimization objective for gradient
    descent, which must be evaluated thousands or millions of times while learning a model's parameters.  We propose a differentiable and efficient approximation of this metric. By incorporating spatial information into the objective we can use a simpler model than competing methods without sacrificing accuracy, resulting in faster inference speeds and alleviating the need for pre/post-processing. We match (or improve) performance  on several tasks compared to prior state of the art by traditional metrics, and in many cases significantly improve performance by the weighted F-measure. 
\end{abstract}

\section{Introduction}

When optimizing a predictive model it is important that the objective
function not only encode the ideal solution (zero mistakes), but also
quantify the relative severity of mistakes. A common dimension of preference is
the desired tradeoff between precision and recall. One can capture this
tradeoff with a $F_\beta$ metric,
where $\beta$ reflects the relative importance of recall compared to
precision. While this metric can quantify the relative importance of
false positives and false negatives, it cannot capture differing
severity between two false positives, or two false
negatives. One domain where differentiating between such errors becomes important is the prediction of foreground maps, where the output has many desired properties not captured by notions of precision or recall, such as
smoothness, accuracy of boundaries, contiguity of the predicted mask, etc. As a result, two
predictions with the
same number of mistakes, or with the same score on a measure
which treats false positives and false negatives equally (e.g. intersection over
union, IoU), may differ substantially in their perceived spatial
quality. Loss functions derived from per-pixel classification-based surrogates,
such as log-loss are almost universally used in existing work, but fail to
capture both the precision-recall tradeoff and the spatial
sensibilities of this kind.

\begin{figure}
\newcommand{\currentimgwidth}{0.2\linewidth}
\newcommand{\currentspace}{.08em}
\setlength\tabcolsep{\currentspace}
\begin{tabular}{ccccc}
\includegraphics[width=\currentimgwidth]{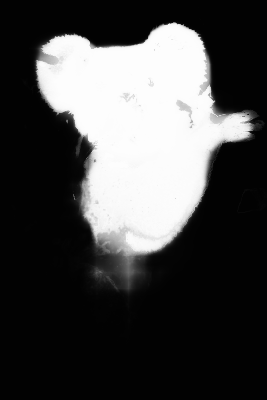}
&
\includegraphics[width=\currentimgwidth]{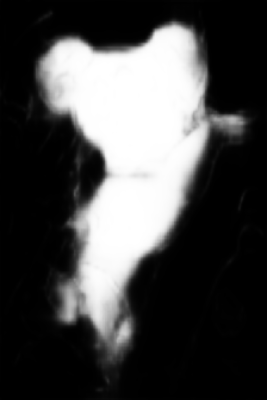}
&
\includegraphics[width=\currentimgwidth]{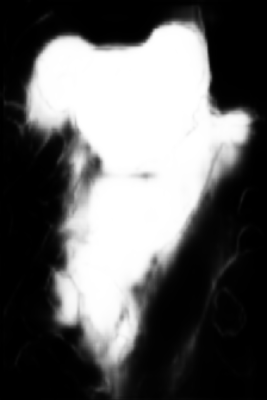}
&
\includegraphics[width=\currentimgwidth]{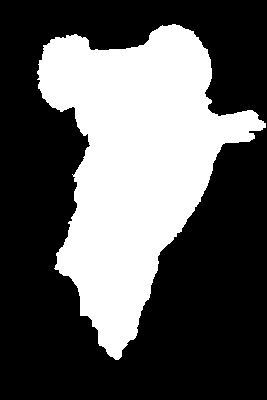}
&
\includegraphics[width=\currentimgwidth]{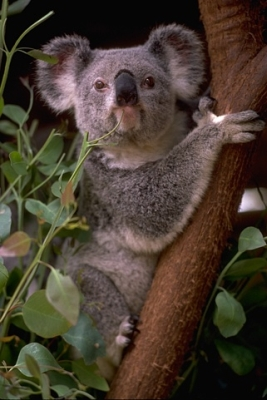} \\
\includegraphics[width=\currentimgwidth]{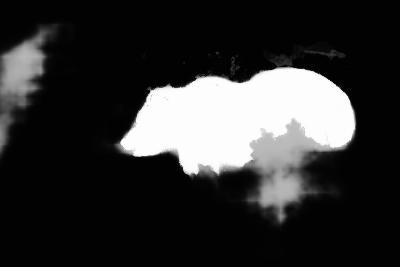}
&
\includegraphics[width=\currentimgwidth]{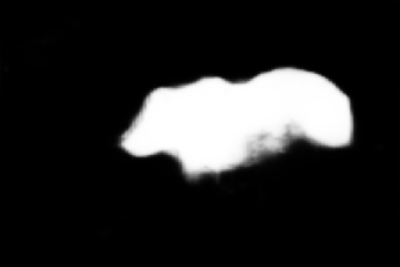}
& \includegraphics[width=\currentimgwidth]{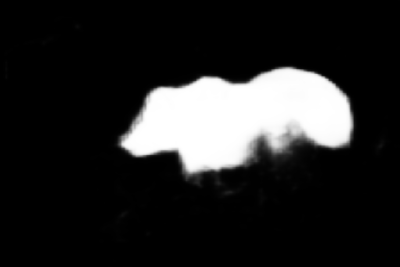}
& \includegraphics[width=\currentimgwidth]{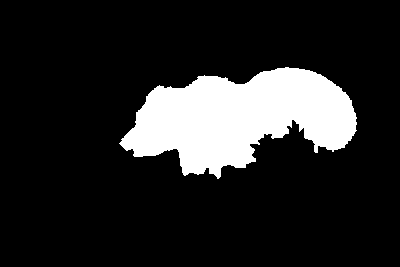}
& \includegraphics[width=\currentimgwidth]{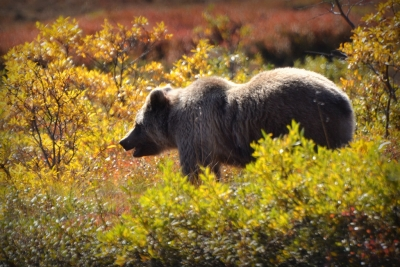} \\
\cite{DCL} & $SZN_{CE}$ & $SZN_{F^w_1}$ & GT & Input \\
 & (ours) & (ours) & & \\
\end{tabular}
\caption{A comparison of the previous saliency prediction state of the art with our $SZN$ model predictions with the traditional log-loss ($SZN_{CE}$) and our proposed $AF^w_\beta$ loss ($SZN_{F^w_1}$). The top row demonstrates that our loss heavily penalizes for large spatially co-occurring false negatives. The bottom row demonstrates that the proposed loss heavily penalizes false positives far from the true object boundary.}
\label{fig:1}
\end{figure}

	Margolin et al.~\cite{margoinEval14} proposed a method to
        quantify these distinctions when predicting foreground
        maps. Their $F_\beta^w$ measure formalizes two notions. First,
        false detections are less severe when close to the object's
        true boundary; Second, missing an
        entire section of an object is worse than missing the same
        number of pixels scattered across the entire object. These
        alterations closely match human intuition and perceptual
        judgements, and have the additional benefits of being robust
        to small annotation errors (such as minor differences between
        multiple human annotators). The $F_\beta^w$ measure is also
        able to reliably rank generic foreground maps, such as
        centered geometric shapes, lower than state of the art
        predictions (They show
        traditional metrics, such as AUC, lack this property). Despite
        these positive traits, their formulation has $O(n^2)$ memory and computational requirements where $n$ the number of pixels in the image.

This computational cost poses a particular problem if the
$F_\beta^w$ metric were used as the training objective for deep neural networks
(DNNs). Normally trained with stochastic gradient descent over large
training sets, DNNs require computing the gradient of the loss many, many,
times. This means that the loss function must be differentiable,
and efficient -- two criteria which
$F_\beta^w$ does not meet.

Our primary contribution is a \emph{differentiable and computationally
efficient} approximation of the $F_\beta^w$ metric, which can be used
directly as the loss function of a convolutional neural network
(CNN). As a secondary contribution, we propose a memory-efficient CNN
architecture which is capable of producing high resolution pixel-wise
predictions, taking full advantage of the spatial information provided
by our proposed loss. By combining these two components we are able to
produce high-fidelity, spatially cohesive predictions, without relying
on complex, often expensive pre-processing (such as super-pixels) or
post-processing (such as CRF inference), resulting in inference speeds an order of magnitude faster than state of the art in multiple domains. We do not sacrifice accuracy, achieving competitive or state of the art
accuracy on benchmarks for salient object detection, portrait
segmentation, and visual distractor masking.

\section{Background}\label{sec:bg}
In this section we discuss the prior work on incorporating
spatial consideration into learning objectives. While multiple objectives have been proposed to
capture spatial properties of prediction
maps~\cite{ahmed2015optimizing,nowozin2014optimal,rosenfeld2014learning},
these have been limited to structured prediction methods using random
fields, and adds significant complexity when incorporated into a feed-forward
prediction framework like that of CNNs.
We focus on the $F_\beta^w$ metric, which is decoupled from the
prediction framework and upon which we directly build our
approach. We review it below, and also survey the related work on the segmentation tasks on which
we evaluate our contributions: salient object detection, distractor
detection, and portrait segmentation.

\subsection{The $F_\beta$ metric family}

In a binary classification scenario, with labels
$y\in\{0,1\}$, when the predicted label $\widehat{y}$ is a mistake
$\widehat{y}\ne y$, it is either a false positive (FP, $\widehat{y}=1$) or a false negative
(FN, $\widehat{y}=0$). Performance of any classifier on an
evaluation set can be characterized by its precision $\#TP/(\#TP+\#FP)$ and
its recall $\#TP/(\#TP+\#FN)$.

While precision and recall each only tell part of
the story, one can summarize a
classification algorithm's performance in a single number, using the
$F_\beta$ metric
\begin{equation}
\label{eq:fbeta}
F_\beta=\frac{(1+\beta^2)*Precision*Recall}{\beta^2Precision+Recall}.
\end{equation}
$\beta$ captures the relative importance of precision compared to recall (e.g. if precision is twice as important as recall, we use $F_2$). The
well known $F_1$ metric is a special case corresponding to equal
importance between precision and recall. The $F_\beta$ metrics is a common 
benchmark in 'information extraction' tasks, and in ~\cite{jansche2005maximum} Jansche
outlines a procedure to directly optimize it. This formulation applies to any scenario when $F_\beta$ is meaningful, but it cannot encode differences \emph{within} the categories of false positive, and false negative, which are quite meaningful in the highly structured domain of natural images.

\subsection{The $F_\beta^w$ metric}~\label{sec:fbeta}

The standard $F_\beta$ is extended in~\cite{margoinEval14} in two
ways. First, it is generalized to handle
continuous predictions, $\widehat{y}\in[0,1]$ (the ground truth $y$
remains binary). The adjusted definitions of the true positive, false
positive, true negative, and false negative are as follows:
\begin{equation}\label{eq:Fbeta}
\begin{split}
E &= |Y-\widehat{Y}| \\
TP&=(1-E)\cdot Y\\
TN&= (1-E)\cdot (1-Y)\\
FP&= E\cdot (1-Y)\\
FN&= E\cdot Y\\
\end{split}
\end{equation}
This holds in the case of predicting a set of values; $Y$ is the vector of ground truth labels , $\widehat{Y}$ is the vector of predictions, and $\cdot$ denotes the dot product

The second modification proposed in~\cite{margoinEval14} addresses the
unequal nature of mistakes in binary segmentation ($y=1$ implying foreground, $y=0$
background), as determined by the spatial configuration of predictions
vs. ground truth. The authors of~\cite{margoinEval14} suggest a number
of criteria for evaluating foreground maps. 

First consider false
negatives, missed detections of foreground pixels. If random foreground
pixels across an object are undetected, leaving small holes in
the foreground, this is easily corrected via
post-processing. However, concentrating the same number of errors in
one part of the object is much more perceptually severe and
difficult to correct. See the
top row of Figure ~\ref{fig:1}. This is captured by by re-weighting
$E\in \mathbb{R}^n$ with a matrix $\mathbb{A} \in \mathbb{R}^{n^2}$:
\begin{equation}
\label{eq:A}
\mathbb{A}=
\begin{cases}
\frac{1}{\sqrt{2\pi\sigma^2}}e^{-\frac{d(i,j)^2}{2\sigma^2}}, & \forall i,j | y_i=1,y_j=1 \\
1, & \forall i,j | y_i=0,i=j\\
0, & \text{otherwise}
\end{cases}
\end{equation}


This definition of $\mathbb{A}$ means that FN error at any given pixel is calculated by summing over all FN errors in the image, weighted by a gaussian centered at the pixel of interest. Intuitively, if there are many spatially co-occurring FN predictions, they will all contribute to each other’s loss, heavily penalizing larger sections of missed foreground. 

False positives, or erroneous foreground detections, are treated differently. A false positive near the true boundary of the object is more acceptable than a distant one. Even human annotators often do not precisely agree on the boundaries of an object. See the bottom row of Figure ~\ref{fig:1}. Margolin et al.~\cite{margoinEval14} quantify this as follows:

   \begin{equation}
   \label{eq:B}
   \mathbb{B}=
   \begin{cases}
   1, & \forall i y_i=1 \\
   2-e^{\alpha\cdot\Delta_i}, & \text{otherwise}
   \end{cases}
   \end{equation}

   \begin{equation}
    \label{eq:AEB}
   	E^w=min(\mathbb{A}E,E)\cdot \mathbb{B}
   \end{equation}

    Where $\Delta_i=\min_{j | y_j=1}d(i,j)$, and $\alpha=\frac{\ln(0.5)}{5}$. Intuitively this gives false positives a weight $\mathbb{B}\in(1,2)^n$, where false positives spatially distant from any true positive approach weight 2, and false positives next to true positives have weight approximately 1. This penalizes more heavily far spurious false detection.

$TP^w,TN^w,FP^w$ and $FN^w$ are then defined by substituting $E^w$ in place of $E$ in Eq.~\ref{eq:Fbeta}.
and use these terms to define weighted precision, weighted recall, and the $F^w_\beta$ metric.
\begin{equation}\label{eq:AFbeta}
\begin{split}
P^w &= \frac{TP^w}{TP^w+FP^w} \\
R^w &= \frac{TP^w}{TP^w+FN^w} \\
F^w_\beta &= \frac{(1+\beta^2)*P^w*R^w}{\beta^2P^w+R^w}
\end{split}
\end{equation}

\begin{figure}[!th]
\newcommand{\currentimgwidth}{0.24\linewidth}
\newcommand{\currentspace}{.08em}
\setlength\tabcolsep{\currentspace}
\begin{tabular}{cccc}
\includegraphics[width=\currentimgwidth]{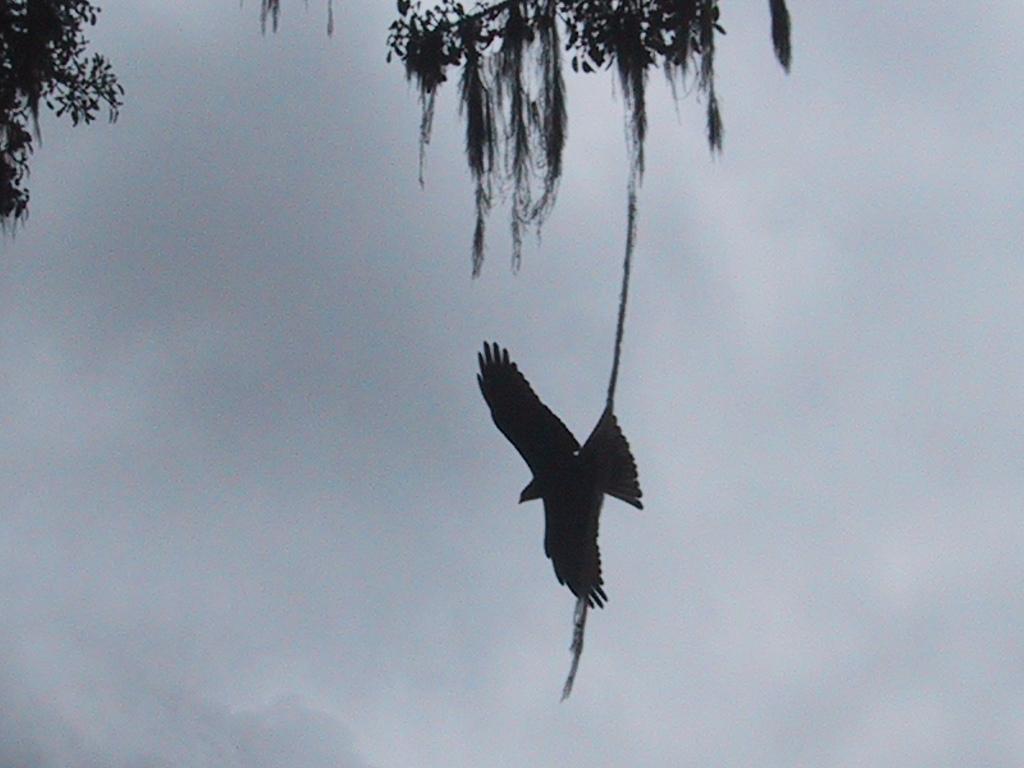}
&
\includegraphics[width=\currentimgwidth]{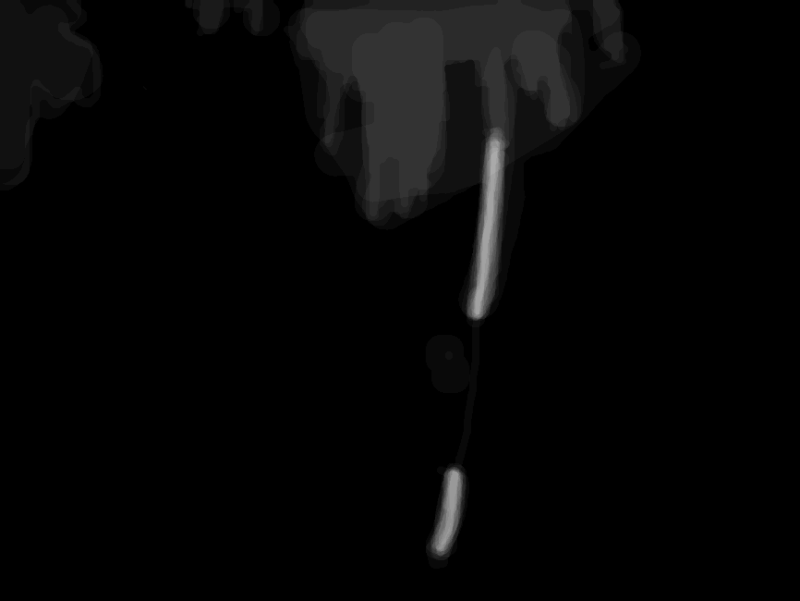}
&
\includegraphics[width=\currentimgwidth]{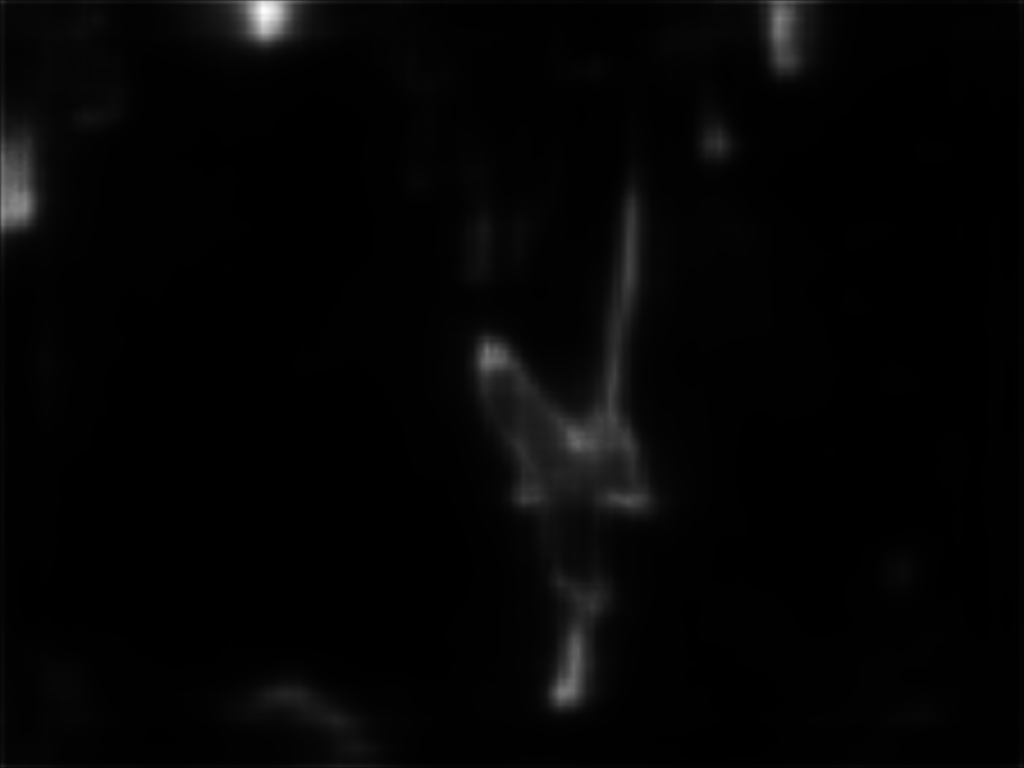}
&
\includegraphics[width=\currentimgwidth]{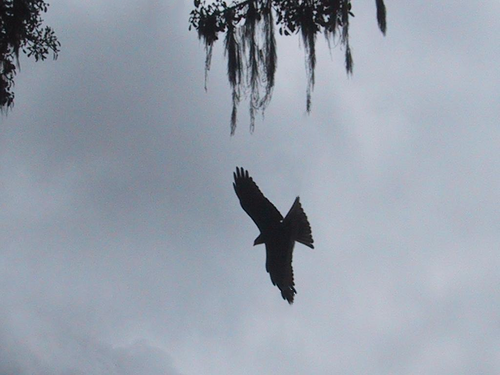} \\
\includegraphics[width=\currentimgwidth]{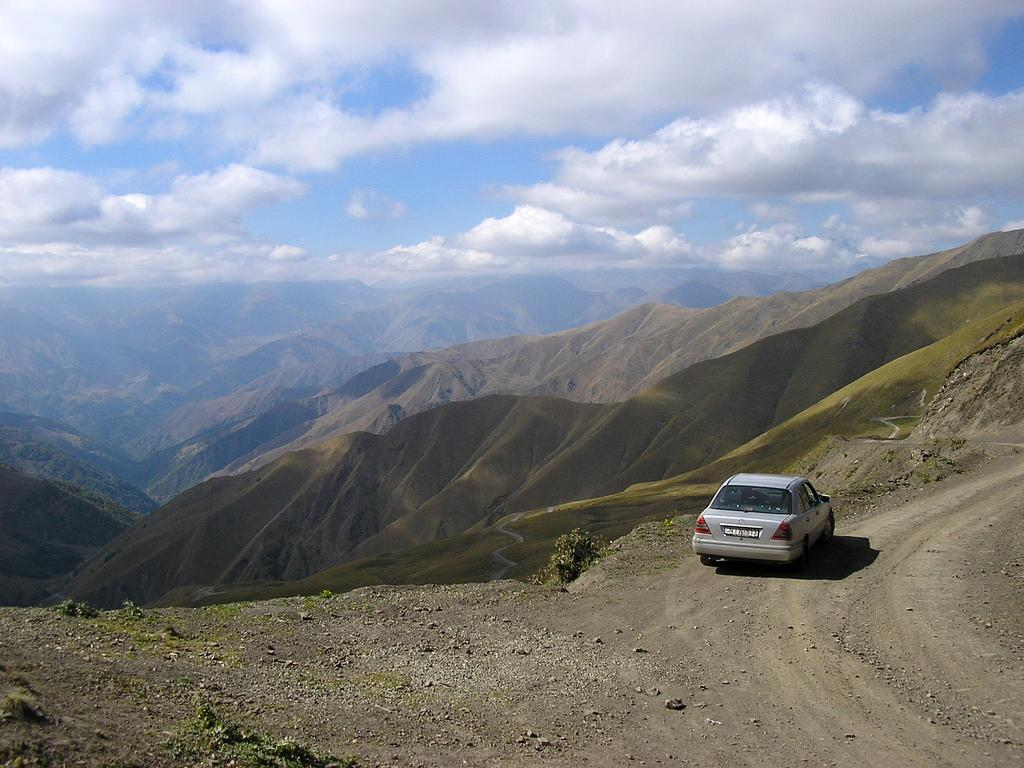}
&
\includegraphics[width=\currentimgwidth]{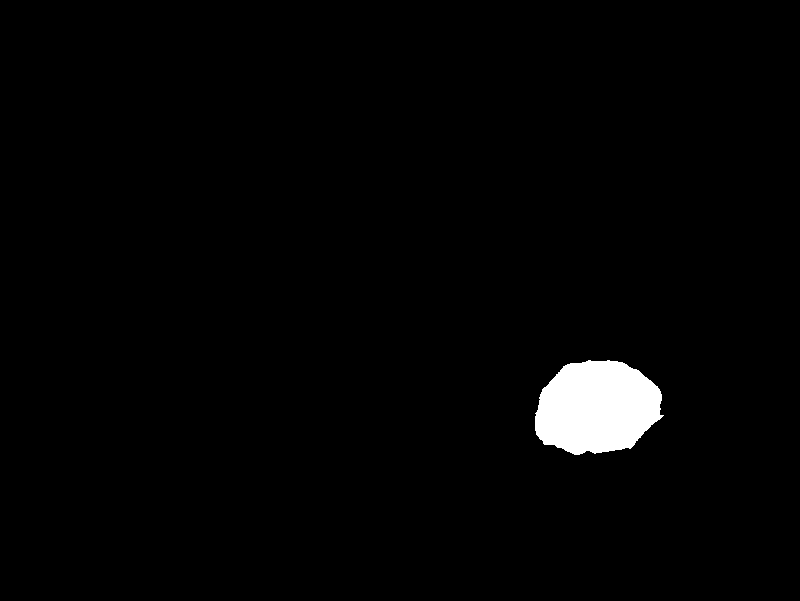}
&
\includegraphics[width=\currentimgwidth]{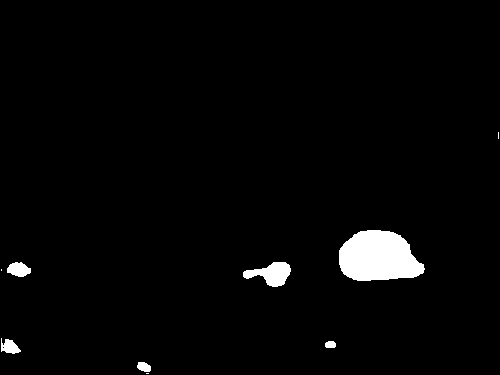}
&
\includegraphics[width=\currentimgwidth]{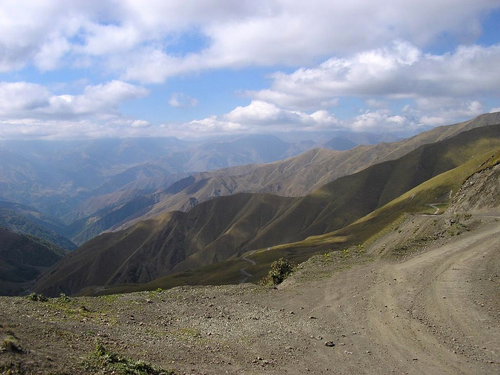} \\
Input & GT & $SZN_{F^w_1}$ & Retouched
\end{tabular}
\caption{Examples of distractor detection and removal (MTurk data
  set, Sec.~\ref{sec:distractors}). Ground truth was obtained by aggregating crowdsourced
  annotations. Our method ($SZN_{F^w_1}$) detects distractors which are then retouched (hole filled) using Photoshop's Content Aware Fill.}
\label{fig:DRem}
\end{figure}

\subsection{Salient Object Detection}
 Traditionally salient object detection models have been constructed by applying expert domain knowledge. Some methods rely on feature engineering combined with center-surround contrast concepts motivated by human perception, where the features are based on color, intensity and texture~\cite{3_viola2001rapid,5_achanta2009frequency,16_cheng2015global}. A more advanced perception model was used in~\cite{27_ma2003contrast} to generate object detections from attended points. 
 Another approach is using high-level object detectors to determine local 'objectness'~\cite{33_judd2009learning}.
 Many methods combine both approaches \cite{31_jia2013category,32_chang2011fusing,33_judd2009learning,29_shen2012unified}. 
 Other techniques make predictions hierarchically ~\cite{15_yan2013hierarchical}, or based on graphical models ~\cite{31_jia2013category,4_liu2011learning}. Other expert knowledge includes re-weighting the model predictions based on the image center or boundaries~\cite{4_liu2011learning,33_judd2009learning,14_jiang2013salient,35_zhu2014saliency}.

Deep networks were used in~\cite{40_wang2015deep} to learn local patch features to predict the saliency score at the center of the patch. However, lack of global information might lead to failure to detect the interior of large objects. 
In ~\cite{kokkinos2016ubernet} Kokkinos combines the task of salient
object detection with several other vision tasks, 
demonstrating a general multi-task CNN architecture.

CNNs were used to extract features around super-pixels~\cite{li2016visual,42_zhao2015saliency}, as well as combining them with hand-crafted ones~\cite{li2016visual}.
Li et al.~\cite{DCL} propose a two stream method that fuses coarse pixel-level prediction, based on concatenated multi-layer features similar to~\cite{mostajabi2015feedforward}, and then fusing these with super-pixel predictions (reminiscent to~\cite{li2016visual}). The results of~\cite{DCL} and~\cite{li2016visual} also rely on post-processing with a CRF.

Our method differs from~\cite{li2016visual,DCL} in two important ways. Instead of relying upon spatial supervision provided by super-pixel algorithms, our architecture {\em directly} produces a high resolution prediction. Our proposed spatially sensitive loss function encourages the learned network to make predictions that snap to object boundaries and avoid ``holes'' in the interior of objects, {\em without any post-processing} (e.g., CRF). Our model achieves competitive or state-of-the-art results on all benchmarks. Additionally, training our model is three times faster than competitive saliency methods, making it much easier to scale to larger training sets. Performing inference with our model is almost an {\em order of magnitude faster} than any competitive method,  and can be used in a real-time application.

\subsection{Distractor Detection}
	Another task where it is vital to predict accurate and high resolution foreground maps is distractor detection as proposed by ~\cite{fried2015finding}. Distractors are defined as visually salient parts of an image which are not the photographer's intended focus. This task is somewhat similar to salient object detection, but successful algorithms must go beyond simply detecting all salient objects, and model the image at a global level to discriminate between the intended focus of the image and the distractors. In ~\cite{fried2015finding} Fried et al. propose an SVM based approach, trained on a relatively small dataset, which classifies super-pixels extracted by Multiscale Combinatorial Grouping (MCG) based on a set of hand-crafted features. To test the robustness of their approach we gathered a larger dataset with crowdsourced labels. While their method is able to detect large and well defined distractors, it struggles to detect non-object distractors such as shadows, lights and reflections as well as select small objects.

\subsection{Portrait segmentation}
Portraits are highly popular art form in both photography and painting. In most instances, artists seek to make the subject stand out from its surrounding, for instance, by making it brighter or sharper or by applying photographic or painterly filters that adapt to the semantics of the image.
Shen et al.~\cite{shen2016portraits} presented a new high quality automatic portrait segmentation algorithm by adapting the FCN-8s framework~\cite{long2014FCN}. They also introduced a portrait image segmentation dataset and benchmark
for training and testing.


\section{Our Approximate $F^w_\beta$ loss ($AF^w_\beta$)}
\begin{figure}
\newcommand{\currentimgwidth}{0.19\linewidth}
\newcommand{\currentspace}{.1em}
\setlength\tabcolsep{\currentspace}

\begin{tabular}{ccccc}
\multirow{3}{*}{\includegraphics[width=\currentimgwidth]{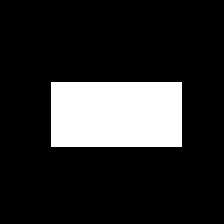}}
&
\includegraphics[width=\currentimgwidth]{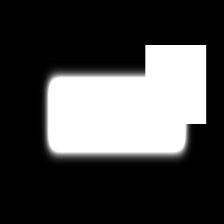}
&
\includegraphics[width=\currentimgwidth]{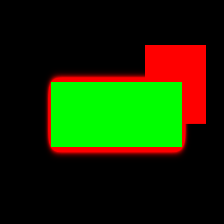}
&
\includegraphics[width=\currentimgwidth]{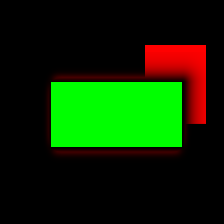}\\
 & Over-Prediction & $E$& $E \mathbb{B}$ \\
&
\includegraphics[width=\currentimgwidth]{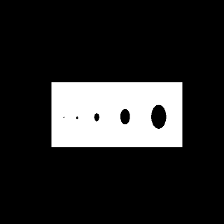}
&
\includegraphics[width=\currentimgwidth]{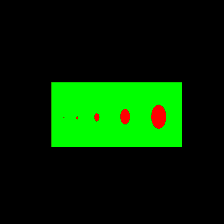}
& \includegraphics[width=\currentimgwidth]{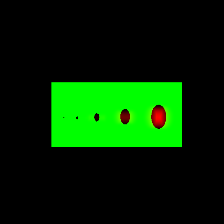}\\
GT & Under-Prediction &  $E$ & $E \mathbb{A}$ \\
\end{tabular}
\caption{A vizualization on a synthetic example of how our loss function re-weights mistakes. In the top row we visualize the down-weighting of false positives near the true object border. In the bottom-row we show the down-weighting of false-negatives do not spatially co-occur many other false-negatives and the increased weight of false negatives which spatially co-occur. True positives are marked by green, and mistakes are marked in red.}
\label{fig:approxQual}
\end{figure}

There are three issues that prevent the $F_\beta^w$ metric as defined in Section \ref{sec:fbeta} from being directly optimized as a loss function. The first is that while the metric is differentiable almost everywhere, it is not differentiable when $y=\hat{y}$, because $\frac{d}{d\hat{y}_i}|\hat{y}_i-y_i|$ is undefined.
\ignorethis{In practice we observed difficulties optimizing the error when it was formulated in this way, and we hypothesize that this is a result of the definition of the gradient of absolute value:
\begin{align*}
\frac{dE}{d\hat{y}_i}|\hat{y}_i-y_i|=
\begin{cases}
1, & \text{ if }\hat{y}>y\\
-1, & \text{ if }\hat{y}<y\\
undefined, & \text{otherwise}
\end{cases}
\end{align*}
Because the gradient remains constant, so long as the prediction is not perfect, we suspect that SGD based algorithms will have a great deal of difficulty converging (intuitively, because the gradient doesn't decrease as the error decreases).}
In practice, we observed difficulties optimizing the error using SGD due to the constant value of the gradient for $\hat{y} \neq y$ (intuitively, because the gradient doesn't decrease as the error decreases). We solve this by replacing the $L_1$ norm with $L_2$: $E_i=(\hat{y}_i-y_i)^2$, which we find to be much easier to optimize.

The second problem is that constructing $\mathbb{A}$ (not to mention computing $E\mathbb{A}$) has $O(n^2)$ time and space complexity. However, we can overcome this problem by leveraging convolutions. When we unpack the definition of matrix multiplication in \eqn{eq:A}, we can write $E\mathbb{A}$ at pixel $i$ as:
    \begin{equation}
    \label{eq:EAunpack}
    \begin{split}
    (E\mathbb{A})_i &= \sum_j[E_j\mathbb{A}_{ji}] = \\
    &= y_i\sum_j[y_jE_j\frac{1}{\sqrt{2\pi\sigma^2}}e^{\frac{-d(i,j)}{2\sigma^2}}]+(1-y_i)
    \end{split}
    \end{equation}

    \ignorethis{
    \begin{equation}
    \label{eq:EAunpack}
    (E\mathbb{A})_i&=\sum_j[E_j\mathbb{A}_{ji}]\\
    &=
    \sum_j[E_j\cdot\begin{cases}
      \frac{1}{\sqrt{2\pi\sigma^2}}e^{-\frac{d(i,j)^2}{2\sigma^2}}, & \forall i,j | y_i=1,y_j=1 \\
      1, & \forall i,j | y_i=0,i=j\\
      0, & \text{otherwise}
    \end{cases}]\\
    &=y_i\sum_j[y_jE_j\frac{1}{\sqrt{2\pi\sigma^2}}e^{\frac{-d(i,j)}{2\sigma^2}}]+(1-y_i)
    \end{equation}
    }

    Note that if $d(i,j) \ge 4\sigma$ then $\frac{1}{\sqrt{2\pi\sigma^2}}e^{\frac{-d(i,j)}{2\sigma^2}}\approx 0$. We then define $y_{p,q}$ and $E_{p,q}$ as the ground truth and error respectively at pixel $(p,q)$, and can approximate $(E\mathbb{A})_i$ as:
    \begin{equation}
    \label{eq:EAfoursig}
    \begin{split}
    (E\mathbb{A})_i &\approx (1-y_i) + \\
    &+ y_{i+(p,q)}\sum_{p,q=-4\sigma}^{4\sigma}[y\cdot E]_{i+(p,q)}\frac{1}{\sqrt{2\pi\sigma^2}}e^{\frac{-\sqrt{p^2 +q^2}}{2\sigma^2}}
    \end{split}
    \end{equation}

If we let $\sigma=\frac{\theta}{4}$, we can define a $(2\theta+1)
\times (2\theta+1)$ Gaussian convolutional kernel $K^{\mathbb{A}}$, yielding

  \begin{equation}
    \label{eq:EAconv}
    \begin{split}
    E\mathbb{A} &\approx Y\odot ((Y\odot E) \ast K^{\mathbb{A}})+(1-Y) \textnormal{, where} \\
    K^{\mathbb{A}}_{i,j}&=\frac{1}{\sqrt{2\pi\sigma^2}}e^{\frac{-\sqrt{(-\theta+i)^2+(-\theta+j)^2}}{2\sigma^2}}
    \end{split}
  \end{equation}

Where $\odot$ is element-wise multiplication. Now we don't need to store any entries of $\mathbb{A}$, only a$(2\theta+1) \times (2\theta+1)$ kernel and we skip most of the original summation over pixels $j$. So the time and space complexity is reduced to $O(n+\theta^2)$. In practice, we use $\theta=9$.

The final problem is that constructing $\mathbb{B}$ has complexity $O(n^2)$, because computing $\Delta_i$ requires finding the minimum $d(i,j)$ over all pixels $j$, such that $y_j=1$. However, if $d(i,j)\ge 25$, then $\mathbb{B}_i=2-e^{\alpha\cdot\Delta_i} \approx 2$ in \eqn{eq:B}. Our intuition is that $\mathbb{B}$ is modeling the region of uncertainty about an object's boundary, for which we believe 25 pixels to be too generous. So we redefine $\Delta_i$ as:
    \begin{align*}
        \Delta_i=
        \begin{cases}
          \infty, & \text{ if }\min_j d(i,j) > 5 \\
          \min_j d(i,j)^2|y_j=1, & \text{ otherwise}
        \end{cases}
    \end{align*}
Squaring the distance so that the region of uncertainty is assumed to be approximately 5 pixels instead of 25. Now the time complexity is $O(n+\phi^2)$ where $\phi=5$. We can approximate $E \odot \mathbb{B}$ using a convolution with the kernel $K^\mathbb{B}$, a tensor of size $\phi \times \phi \times \phi^2$, defined at each index (zero indexed) as:
    \begin{align*}
        K^\mathbb{B}_{i,j,k}=
        \begin{cases}
          (-\phi+i)^2+(-\phi+j)^2, & \text{ if }k=i*(2\phi+1)+j \\
          0, & \text{otherwise}
        \end{cases}
    \end{align*}
and we can rewrite $\mathbb{B}$ as:

  \begin{equation}
    \label{eq:Bapprox}
    \mathbb{B} \approx Y+ (1-Y)\odot e^{a\min_k[(Y \ast K^\mathbb{B})]}
  \end{equation}

By reformulating the local search for the true object boundary as a
convolution followed by an argmin, we can leverage the efficient implmentations of
these operations already available in many packages. While our current architecture does not suffer from speed or memory issues, more complicated architectures might benefit from a more optimized implementation, namely a custom 'minimization convolution' that would not store the intermediary result of $(Y \ast K^\mathbb{B})$, and takes advantage of the sparsity of $K^\mathbb{B}$.

These changes yield a spatially informed loss function that can easily be implemented in an existing DNN framework such as Tensorflow. It fully utilizes the GPU, does not increase training wall clock time noticeably, and yields better results than more commonly used loss functions for foreground maps. Compared to an unoptimized implementation of the original formulation in python, our approximation takes two orders of magnitude less time to compute on the CPU, and three orders of magnitude less time on the GPU, to compute our loss on a 224x224 pixel image, See section 5.5.

\section{Network Architecture}\label{sec:arch}
In order to produce accurate foreground maps each pixel must have a rich feature representation. To achieve this we utilize Zoomout features ~\cite{mostajabi2015feedforward}, which have been effectively utilized in the semantic segmentation community. Zoomout features are extracted from a CNN by upsampling and downsampling the features computed by each convolutional filter to be the same spatial resolution, then concatenating the features computed at all layers of the CNN. In this way, each spatial location is richly described by both the weakly localized semantic features computed at higher layers, the strongly localized edge and color detectors computed in the first layers, and everything in between.

\subsection{Squeeze Layers}
Zoomout features are expressive but have a large memory footprint, limiting the spatial resolution of predictions that can be made using them. In tasks like distractor detection, where the end goal is to precisely localize distractors and remove them, a low resolution prediction leads to spatial ambiguity and lower precision. To remedy this problem we adapt the insights of~\cite{iandola2016squeezenet}, introducing what we call \textit{Squeeze Modules} to our network. A Squeeze Module consists of $2n$ convolutional filters, $n$ of which are $1 \times 1$ convolutions and $n$ of which are $3 \times 3$ convolutions. Applying Squeeze Modules to each convolutional layer acts as a dimensionality reduction with learned parameters, allowing us to make predictions at essentially arbitrary resolutions by setting $n$ to be sufficiently small. In practice we produce $224 \times 224$ predictions, and set $n=64$. We refer to our full architecture as a {\em Squeezed Zoomout Network ($SZN$)}.

\begin{figure}[t]
\begin{center}
\includegraphics[scale=0.35]{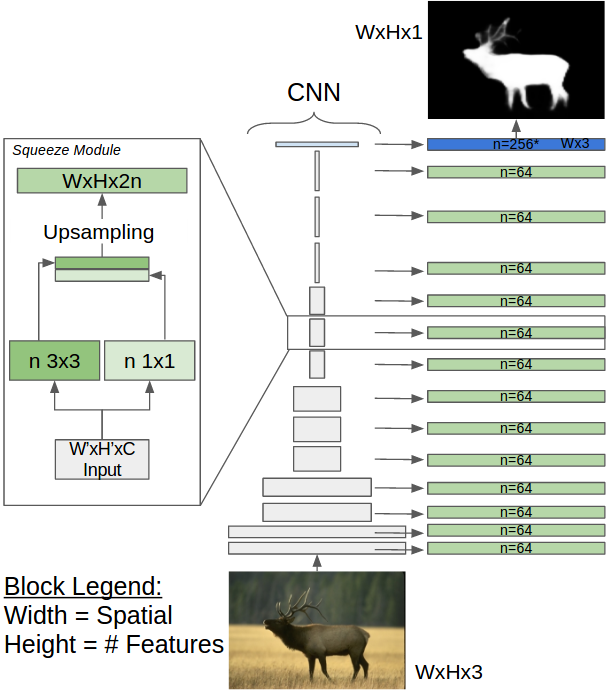}
\end{center}
\caption{Architecture Diagram, note that because the blue squeeze module is applied to a fully connected layer, it uses only 1x1 convolutions.}
\end{figure}


\section{Experiments}
We report on experiments with three tasks where we can expect spatial
sensitivity to be important for quality of the output: salient object
detection, portrait segmentation, and distractor
detection. See Sec.~\ref{sec:bg} for background.

\subsection{Training Details}
In all experiments we train our $SZN$ using a CNN (from which the
squeezed zoomout features are derived) pre-trained on ImageNet. As the
base CNN we use VGG-16~\cite{simonyan2014very} for saliency, portraits, and distractors.
We train the $SZN$ architecture using ADAM~\cite{Kingma14ADAM}, and
train in 3 stages. In the first stage we set the learning rate to 3e-4
for 8 epochs, In the second we set the learning rate to be 1e-4 for 4
epochs. The base CNN is kept fixed (not fine-tuned) in the first two
stages. In the third stage we set the learning rate to 1e-5 for 14
epochs, fine-tuning the weights of the base CNN as well. We augment the training images by randomly permuting standard data transformations as described in~\cite{DFIB15}: image flips, random noise, changing contrast levels, and global color shifts.

\subsection{Salient Object Detection}

We consider four standard data sets for this task:

\textbf{MSRA-B} - 5000 images with pixel level annotations provided by ~\cite{14_jiang2013salient}. Widely used for salient object detection. Most images contain a single object on a high contrast background.

\textbf{HKU-IS} - 4447 images with pixel level annotations provided by ~\cite{li2016visual}. All images with at least one of the following attributes: multiple salient objects, salient objects touching boundary, low color contrast, complex background.

\textbf{ECSSD} - 1000 challenging images with pixel level annotations
provided by ~\cite{15_yan2013hierarchical}.



\textbf{PASCAL-S} 850 images from the PASCAL VOC 2010 segmentation challenge with pixel level annotations provided by ~\cite{49_li2014secrets}. Following the convention of ~\cite{49_li2014secrets} we threshold the soft labels at 0.5.

\begin{figure*}
\newcommand{\currentimgwidth}{.14\linewidth}
\newcommand{\currentspace}{.1em}
\setlength\tabcolsep{\currentspace}

\begin{tabular}{ccccccc}
\includegraphics[width=\currentimgwidth]{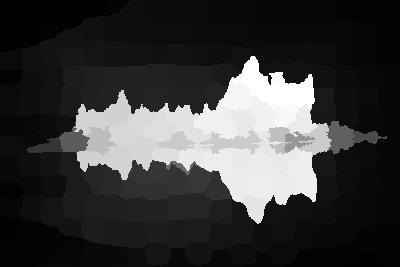}
&
\includegraphics[width=\currentimgwidth]{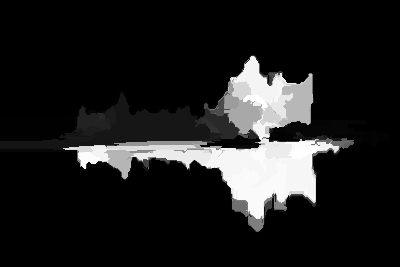}
&
\includegraphics[width=\currentimgwidth]{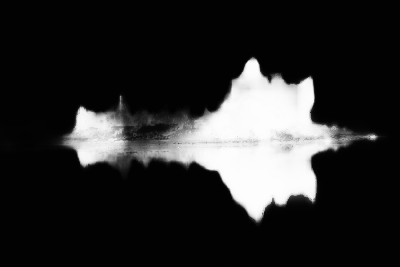}
&
\includegraphics[width=\currentimgwidth]{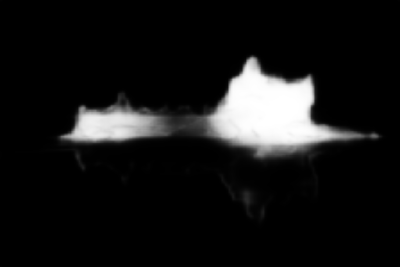}
&
\includegraphics[width=\currentimgwidth]{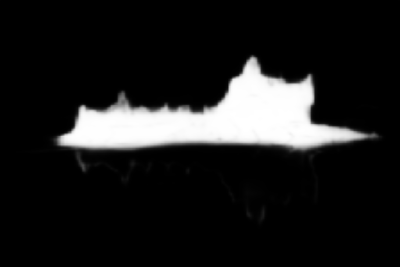}
&
\includegraphics[width=\currentimgwidth]{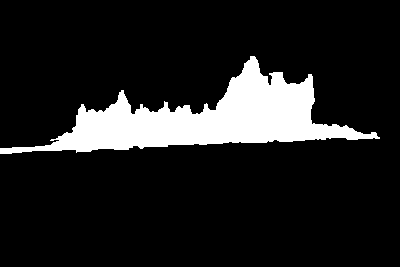}
&
\includegraphics[width=\currentimgwidth]{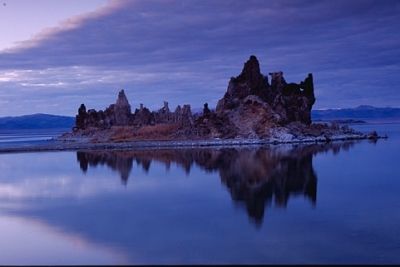}\\

\includegraphics[width=\currentimgwidth]{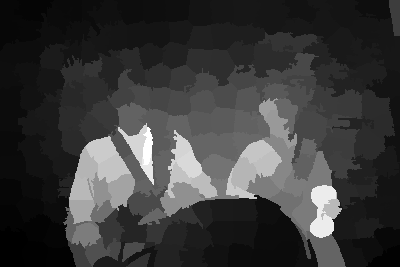}
&
\includegraphics[width=\currentimgwidth]{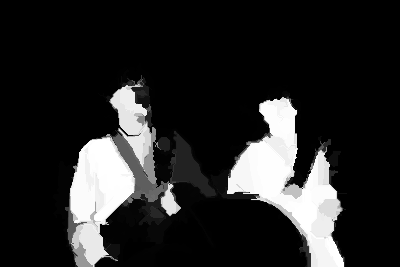}
&
\includegraphics[width=\currentimgwidth]{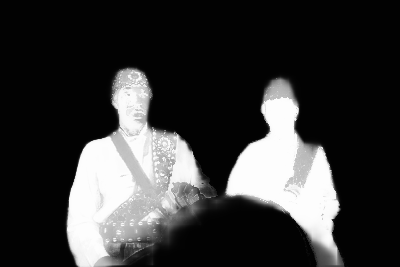}
&
\includegraphics[width=\currentimgwidth]{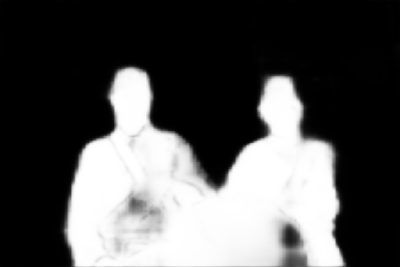}
&
\includegraphics[width=\currentimgwidth]{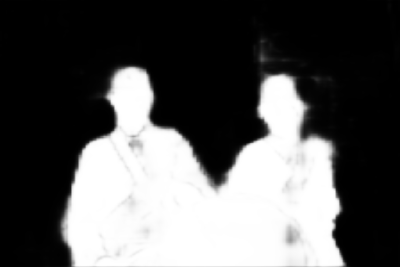}
&
\includegraphics[width=\currentimgwidth]{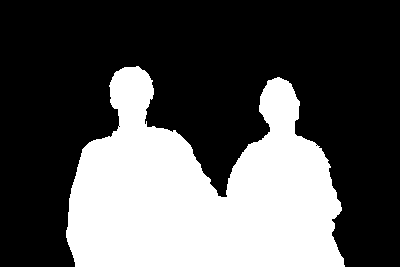}
&
\includegraphics[width=\currentimgwidth]{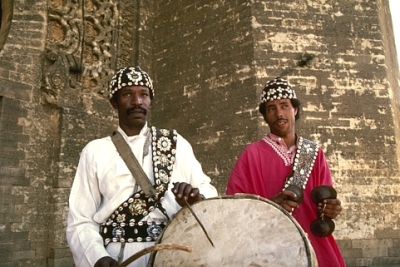}\\

\includegraphics[width=\currentimgwidth]{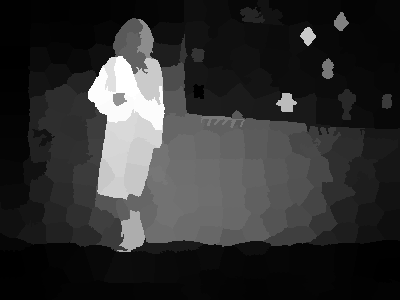}
&
\includegraphics[width=\currentimgwidth]{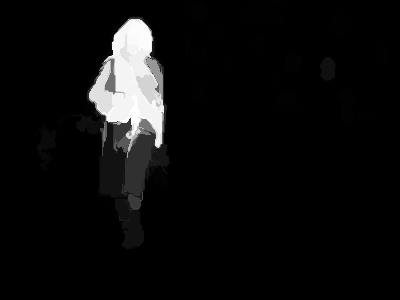}
&
\includegraphics[width=\currentimgwidth]{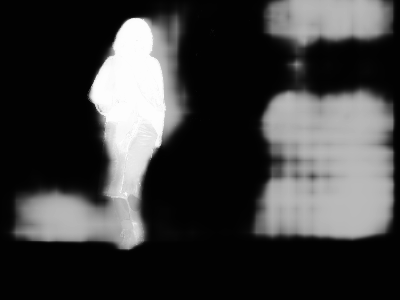}
&
\includegraphics[width=\currentimgwidth]{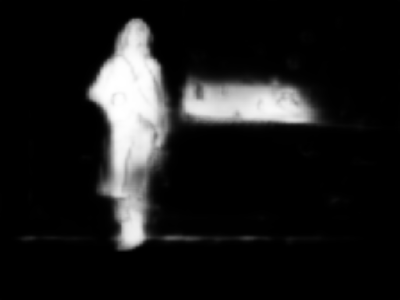}
&
\includegraphics[width=\currentimgwidth]{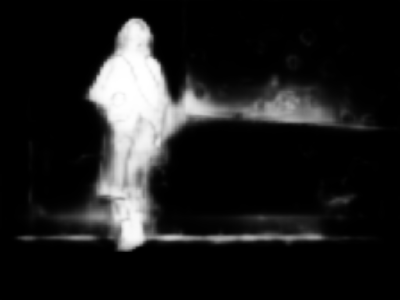}
&
\includegraphics[width=\currentimgwidth]{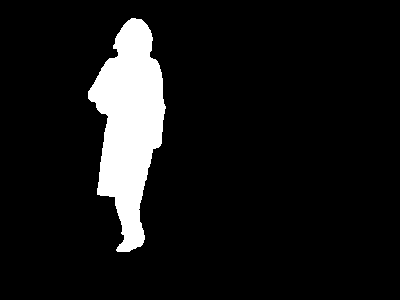}
&
\includegraphics[width=\currentimgwidth]{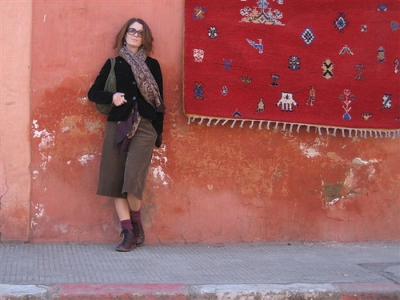}\\

\includegraphics[width=\currentimgwidth]{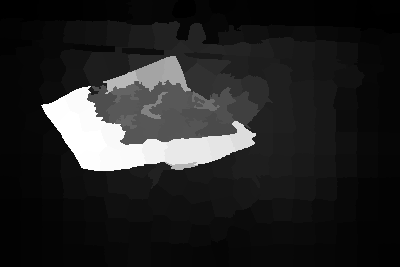}
&
\includegraphics[width=\currentimgwidth]{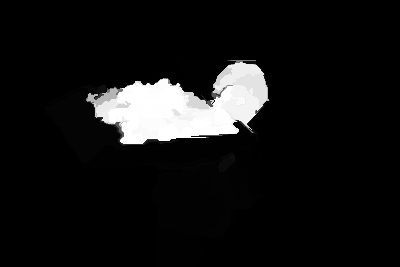}
&
\includegraphics[width=\currentimgwidth]{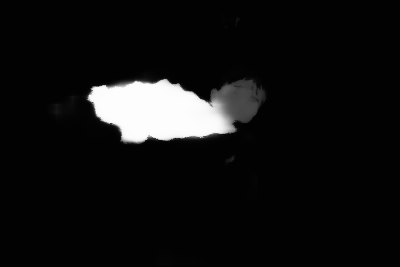}
&
\includegraphics[width=\currentimgwidth]{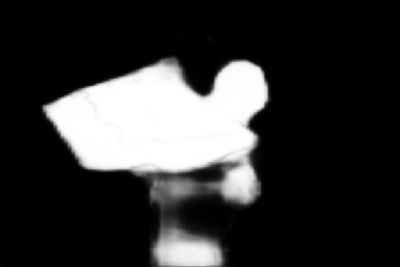}
&
\includegraphics[width=\currentimgwidth]{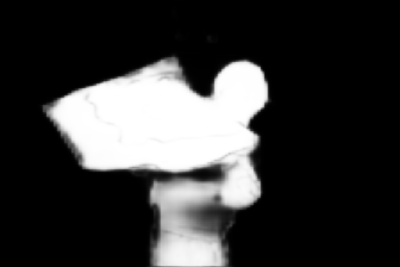}
&
\includegraphics[width=\currentimgwidth]{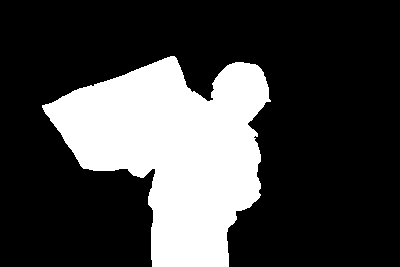}
&
\includegraphics[width=\currentimgwidth]{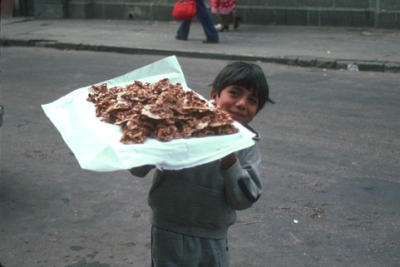}\\



MC~\cite{42_zhao2015saliency} & MDF~\cite{li2016visual} & DCL~\cite{DCL} & $SZN_{CE}$ (our) & $SZN_{F^w_1}$ (our) & GT & Input
\end{tabular}
\vspace{-.5em}\caption{A qualitative comparison of our method with other leading
  methods on object saliency. $SZN_{CE}$: our network trained with
  cross-entropy loss; $SZN_{F^w_1}$: our network trained with the proposed $AF^w_\beta$.}
\label{fig:segQual}
\vspace{1cm}
\end{figure*}

\newcommand{\sameIouLine}[1]{
  \includegraphics[width=\currentimgwidth]{sameIOU_xe_#1}
&
\includegraphics[width=\currentimgwidth]{sameIOU_wf_#1}
&
\includegraphics[width=\currentimgwidth]{sameIOU_gt_#1}
&
\includegraphics[width=\currentimgwidth]{sameIOU_im_#1}}

\newcommand{\sameIouLinePor}[1]{
  \includegraphics[width=\currentimgwidth]{sameIOUPor_xe_#1}
&
\includegraphics[width=\currentimgwidth]{sameIOUPor_wf_#1}
&
\includegraphics[width=\currentimgwidth]{sameIOUPor_gt_#1}
&
\includegraphics[width=\currentimgwidth]{sameIOUPor_im_#1}}

\begin{figure*}
\newcommand{\currentimgwidth}{.12\linewidth}
\newcommand{\currentspace}{.1em}
\setlength\tabcolsep{\currentspace}

\begin{tabular}{cccccccc}
\sameIouLine{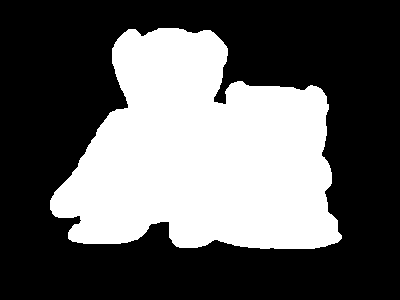}&
\sameIouLine{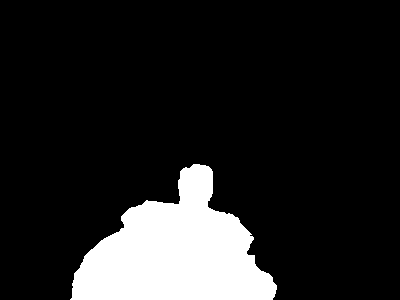}\\
\sameIouLine{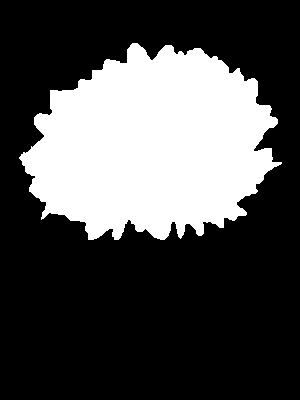}&
\sameIouLine{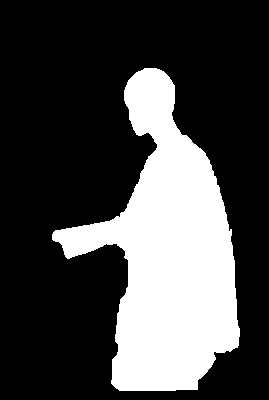}\\
\sameIouLine{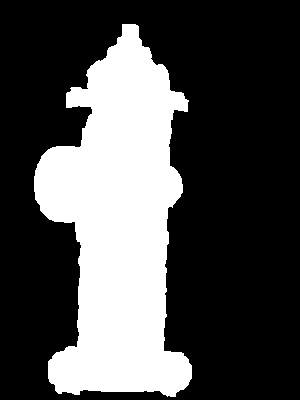}&
\sameIouLine{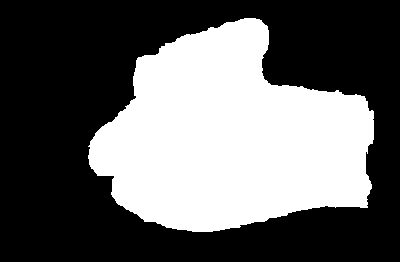}\\

$SZN_{CE}$ & $SZN_{F^w_1}$ & GT & Image&
$SZN_{CE}$ & $SZN_{F^w_1}$ & GT & Image
\end{tabular}
\caption{A visualization of the perceptual importance of the
  $F^w_\beta$ metric on object saliency. On each image, after
  thresholding prediction maps at 0.5, there is a less than 5\%
  difference in the IOU score of the outputs of $SZN_{CE}$ and $SZN_{F^w_1}$, but at
  least a 20\% percent difference in their $F^w_\beta$
  score. Artifacts present in $SZN_{CE}$ outputs but alleviated in $SZN_{F^w_1}$
  outputs include large interior holes, isolated blobs, and poorly
  defined outlines.}
\label{fig:sameIOU}
\vspace{1cm}
\end{figure*}
\noindent\textbf{Evaluation Metrics}

Following the convention of ~\cite{DCL}~\cite{li2016visual} we report the $F_{0.3}$ measure (with oracle access to the optimal threshold for the soft predictions), area under the receiver-operator characteristic curve (AUROC), and mean absolute error ($MAE =\frac{1}{W\cdot H}\sum_{i=1}^w\sum_{j=1}^H |\hat{Y}_{ij}-Y_{ij}|$). While the first two metrics evaluate whether we rank pixels correctly, MAE captures absolute classification error. We report the mean of each metrics on the test set. We also report the $F^w_1$ metric almost exactly as formulated by ~\cite{margoinEval14}, except that for tractibility we drop terms tied to spatially distant pixels, which are very expensive to all compute and have a negligable effect on the loss. While other measures give all errors equal weight, and a small percentage of pixels predicted differently barely affects their value, those mislabeled pixels can be perceptually vital. This is captured by the $F^w_1$ metric, we provide examples of this phenomenon in Figure ~\ref{fig:sameIOU}

\noindent\textbf{Results and Comparison}
Following the convention of ~\cite{DCL}~\cite{li2016visual} we train on 2500 images from MSRA-B, validate on 500, and test on the remaining 2000. We then use the same model trained on MSRA-B to generate predictions for all other datasets.
To evaluate our proposed loss function we compare the performance of
our Squeezed Zoomout Network trained with the commonly used
cross-entropy loss function ($SZN_{CE}$), against the same architecture
trained with the exact same training procedure, but replacing the
cross entropy with our $AF^w_\beta$ loss function. The latter is our
proposed method and we denote it {\em $SZN_{F^w_1}$} from now on. We also
compare both these models against other competitive techniques, MC
~\cite{42_zhao2015saliency}, HDHF ~\cite{li2016visual}, and DCL
~\cite{DCL}, these results are summarized in Table ~\ref{fig:segQuan},
and we provide a qualitative comparison in Figure ~\ref{fig:segQual}
. While ~\cite{kokkinos2016ubernet},~\cite{xi2017fast},and
~\cite{chandra2016deep} report competitive results on some of the same
test sets, they train on 10,000 images, while we only train on 2500,
making the results not directly comparable, and we omit those methods
from~\ref{fig:segQuan}.

We also use saliency to explore the effectiveness of the proposed objective function, compared to other reweighting schemes. These include: Dropping either the reweighting by the matrix $A$ or $B$, using a weighted Cross-Entropy loss, with double weight given to correctly classifying the foreground or background, and standard cross entropy, but ignoring the labels of all pixels in a 3-pixel band around the borders of the foreground. Each of these reweighting schemes reduces AUROC by close to 1\%, but effect on $F^w_1$ varies. Most interesting is the large drop in performance caused by ignoring a 3-pixel border during training, which seems to indicate that these border pixels contain extremely important information for learning a higher quality model. 


All inference timing results were gathered using a Titan X GPU and a 3.5GHz Intel Processor. For training  MC ~\cite{42_zhao2015saliency} used a Titan GPU and a 3.6GHz Intel Processor, HDHF ~\cite{li2016visual} and DCL~\cite{DCL} both use a Titan Black GPU and a 3.4GHz Intel Processor. 


\begin{table}[t]
{\small
\begin{center}
\setlength\tabcolsep{2.4pt}
\begin{tabular}{ |p{4.4em}|p{2.8em}|p{2.5em}|>{\centering\arraybackslash}p{3em}|p{2.3em}|>{\centering\arraybackslash}p{3.5em}|>{\centering\arraybackslash}p{3.5em}|}
	\hline
    & & MC & HDHF & DCL & $SZN_{CE}$ & $SZN_{F^w_1}$ \\
    & & ~\cite{42_zhao2015saliency} & ~\cite{li2016visual} & ~\cite{DCL} & ours & ours \\\hline
    \multirow{4}{*}{MSRA-B}&MAE            & 0.054 & 0.053 & \textbf{0.047} & 0.052 & 0.051 \\ \cline{2-7}
    					   &AUC            & 0.975 & 0.982  & 0.983  & 0.987  &\textbf{0.988} \\ \cline{2-7}
                           &$F_{0.3}^{max}$& 0.984 & 0.899 & 0.916 & 0.913 &\textbf{0.919} \\ \cline{2-7}
                           &\cellcolor{blue!25}$F^w_1$            &\cellcolor{blue!25} - &\cellcolor{blue!25} -  &\cellcolor{blue!25} 0.816   &\cellcolor{blue!25} 0.829  &\cellcolor{blue!25} \textbf{0.856} \\ \hline
    \multirow{4}{*}{HKU-IS}&MAE            & 0.102 & 0.066  & \textbf{0.049} & 0.057 &0.057 \\ \cline{2-7}
    					   &AUC            & 0.928 & 0.972  & 0.981   & 0.985  &\textbf{0.987} \\ \cline{2-7}
                           &$F_{0.3}^{max}$& 0.798 & 0.878  & \textbf{0.904} & 0.891 & \textbf{0.904} \\ \cline{2-7}
                           &\cellcolor{blue!25}$F^w_1$          &\cellcolor{blue!25} - &\cellcolor{blue!25} -  &\cellcolor{blue!25} 0.768   &\cellcolor{blue!25} 0.788  &\cellcolor{blue!25} \textbf{0.826} \\ \hline
	\multirow{4}{*}{ECSSD}&MAE             & 0.100 & 0.098  & 0.075 & \textbf{0.069} & 0.073 \\ \cline{2-7}
    					   &AUC            & 0.948 & 0.960 & 0.968 & \textbf{0.981} & 0.980 \\ \cline{2-7}
                           &$F_{0.3}^{max}$& 0.837 & 0.856 & \textbf{0.924} & 0.905 & 0.908 \\ \cline{2-7}
                           &\cellcolor{blue!25}$F^w_1$&\cellcolor{blue!25} - &\cellcolor{blue!25} -  &\cellcolor{blue!25} 0.767   &\cellcolor{blue!25} 0.796  &\cellcolor{blue!25} \textbf{0.827} \\ \hline
  \multirow{4}{*}{PASCAL-S}&MAE            & 0.145 & 0.142 & 0.108 & \textbf{0.106} & 0.109 \\ \cline{2-7}
    					   &AUC            & 0.907 & 0.922 & 0.924   & \textbf{0.954}   & \textbf{0.954} \\ \cline{2-7}
                           &$F_{0.3}^{max}$& 0.740 & 0.781  & 0.822 & 0.833 & \textbf{0.839} \\ \cline{2-7}
                           &\cellcolor{blue!25}$F^w_1$            &\cellcolor{blue!25} - &\cellcolor{blue!25} -  &\cellcolor{blue!25} 0.670   &\cellcolor{blue!25} 0.657  &\cellcolor{blue!25} \textbf{0.680} \\ \hline\hline
 \multicolumn{2}{|c|}{Train Speed} & \textit{19H} & \textit{12H} & \textit{15H} & \multicolumn{2}{c|}{\textbf{4 H}} \\ \hline
 \multicolumn{2}{|c|}{Test Speed} & 1.1s & \textit{2.5s}& 0.88s & \multicolumn{2}{c|}{\textbf{0.094s }} \\ \hline
\end{tabular}
\caption{Quantitative comparisons between our approach and other leading methods. MAE and - lower is better; AUC, $F_{0.3}^{max}$, and $F^w_1$ - higher is better. Italics indicate a projected training speedup of 1.67 if run on our hardware}
\label{fig:segQuan}
\end{center}
}
\end{table}

\begin{table*}
\setlength\tabcolsep{2.5pt}
\begin{minipage}[c]{.2\textwidth}
{\small
\begin{tabular}{ |p{3em}|p{2.6em}| p{2.3em} |}
	\hline
   & MTurk & Dist9 \\\hline
  \cite{fried2015finding} & 0.81 & 0.67 \\\hline
  Ours & 0.84 & 0.87 \\\hline
  Human & 0.89 & - \\\hline
\end{tabular}}
\captionof{table}{Comparison with Fried et al.~\cite{fried2015finding}
  on distractor detection.}
\label{fig:DistQuan}
\end{minipage}\hspace{1em}%
\begin{minipage}[c]{.4\textwidth}
\includegraphics[width=\textwidth]{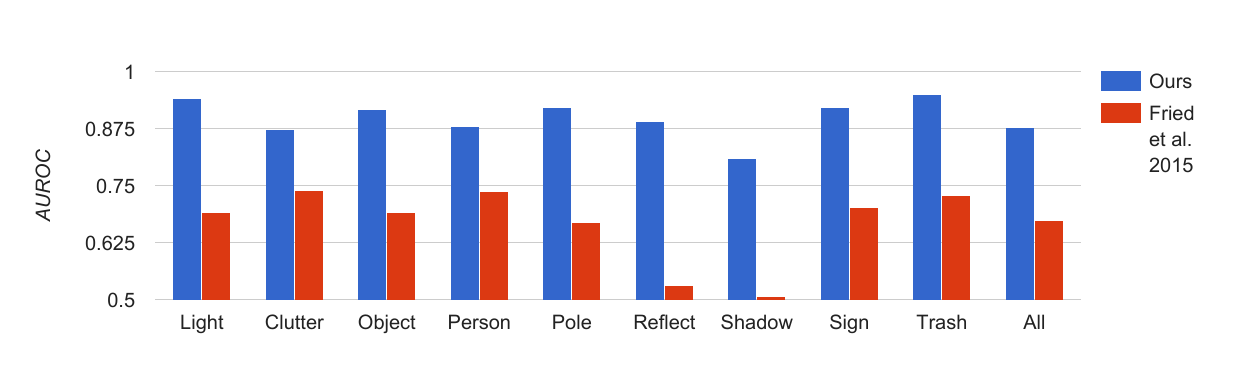}
\captionof{figure}{Distractor detection results across different categories in Dist9 dataset}
\label{fig:DistBar}
\end{minipage}\hspace{1em}%
\begin{minipage}[c]{.35\textwidth}
{\small
\begin{tabular}{|l|c|c|c|}
\hline
             & MIoU & $F^w_1 $ & Test Speed \\
             \hline
$PFCN$~\cite{shen2016portraits}  & 94.20 & 0.965 & 0.114s\\
\hline
$PFCN+$~\cite{shen2016portraits} & 95.91 & 0.972 &  1.125s \\
\hline
$SZN_{CE}$          & 96.53 & 0.965 & \textbf{0.036s} \\ 
\hline
$SZN_{F^w_1}$         & \textbf{97.13} & \textbf{0.973} & \textbf{0.036s} \\ 
\hline
\end{tabular}}
\captionof{table}{Portrait segmentation results. $PortraitFCN+$  ~\cite{shen2016portraits} augments images with 3 extra
  channels, See text for details.}
\label{fig:quanPortraits}
\end{minipage}\hspace{1em}%
\end{table*}

\begin{table}
\begin{tabular}{|l|c|c|}
\hline
             & AUROC & $F^w_1$  \\
             \hline
Proposed  & \textbf{0.988} & \textbf{0.856}\\
\hline
Proposed, no A  & 0.976 & 0.836\\
\hline
Proposed, no B  & 0.975 & 0.835\\
\hline
Cross-Entropy, 2x foreground weight  & 0.976 & 0.834\\
\hline
Cross-Entropy, 2x background weight  & 0.974 & 0.797\\
\hline
Cross-Entropy, 3pix 'DNC' band  & 0.973 & 0.807\\
\hline
\end{tabular}
\captionof{table}{Objective function comparison, see text for details}
\label{tab:ablation}
\end{table}

\subsection{Portrait Segmentation}\label{sec:portraits}
\noindent\textbf{Dataset}
We use the dataset from~\cite{shen2016portraits}, consisting of 1800
human portrait images gathered from
Flickr. A face detector is run on each image, producing a centered crop scaled to be an 800x600. The crop is manually segmented using Photoshop's ``quick select''. This
dataset focuses on portraits captured using a front-facing mobile camera (through the choice of Flickr queries), but includes other portrait types as well. The dataset is split into 1500 training images and 300 test images. There is a wide variety in the subjects' age, clothing, accessories, hair-style, and background.


\noindent\textbf{Results}
Table~\ref{fig:quanPortraits} shows that by MIoU both our models significantly outperform $PFCN$ (PortraitFCN), which uses only RGB input; and $PFCN+$, which requires substantial preprocessing (fiducial point detection, computing an average segmentation mask and aligning it to the input face location) and additional input channels. While our $SZN_{F^w_1}$ model achieves significantly higher $F^w_1$ scores than $PFCN$ and $SZN_{CE}$, they are only slightly better than $PFCN+$. We believe this is due to the spatial guidance used by $PFCN+$.

\subsection{Distractor Detection}\label{sec:distractors}

\noindent\textbf{Datasets}

\textbf{MTurk} - A dataset of 403 images, with accurate, pixel level annotations averaged over many (on average 27.8 ~\cite{fried2015finding}) humans through Mechanical Turk. 

\textbf{Dist9} - A dataset of 4019 images, gathered via a free app which removed regions highlighted by users. Because the ground truth was gathered based on thumb swipes it is often inexact, and has only weak correspondence with object boundaries. To rectify this we used ground truth with scores averaged over super pixels generated with MCG ~\cite{arbelaez2014multiscale}, where the boundary threshold is set to be 0.1. In this dataset each pixel is labeled with either one of 9 foreground classes corresponding to different types of distractors (light, object, person, clutter, pole, trash, sign, shadow, and reflection) or background.

\noindent\textbf{Evaluation}
We evaluate our performance on the MTurk dataset through 10 fold cross validation, and compare against the performance of \cite{fried2015finding} using leave-one-out cross validation. Note that this disadvantages our method, because while each model they use for validation is trained on 402 images, each model we use is trained on 362 or 363 images. We also compare against \cite{fried2015finding} on the Dist9 dataset, training 10 separate models, one on the entire dataset, and one each of the 9 small datasets corresponding to one of the foreground classes. We split the dataset randomly, using 90\% to train and 10\% to test. Following the convention of ~\cite{fried2015finding}, we measure AUROC on all datasets. The results are summarized in Table ~\ref{fig:DistQuan}, and Figure ~\ref{fig:DistBar}. Note the final column in Figure ~\ref{fig:DistBar} averages across categories, while Table ~\ref{fig:DistQuan} averages over the entire dataset.

\subsection{Approximation speed}
To evaluate the relative speed of our approximation we compute wall clock time of computing the $F^w_\beta$, and $AF^w_\beta$ scores, averaged
on fifteen random images from ECSSD. While the original $F^w_\beta$ takes 37 minutes, our approximation takes 8.7 seconds on a cpu, and 0.33 seconds on a GPU.

\section{Discussion and Future Work}
We propose a differentiable and efficient objective function which directly encoding multiple widely desirable spatial properties of a foreground
mask.
We use this objective to learn the parameters of a novel ``squeezed zoomout'' architecture. resulting in high fidelity foreground maps, which
match or surpass state of the art results for a range of
binary segmentation tasks. Notably, we achieve these results without
relying on any pre-processing (e.g., super-pixel segmentation) or
post-processing (e.g., CRF). An interesting direction for fugure work is to
generalize our loss function to a multi-class setting, for instance
semantic segmentation.

\newpage
{\small
\bibliographystyle{ieee}
\bibliography{iccv}
}
\end{document}